\documentclass[letterpaper]{article} % DO NOT CHANGE THIS
\usepackage{aaai2026}  % DO NOT CHANGE THIS
\usepackage{times}  % DO NOT CHANGE THIS
\usepackage{helvet}  % DO NOT CHANGE THIS
\usepackage{courier}  % DO NOT CHANGE THIS
\usepackage[hyphens]{url}  % DO NOT CHANGE THIS
\usepackage{graphicx} % DO NOT CHANGE THIS
\usepackage{natbib}  % DO NOT CHANGE THIS AND DO NOT ADD ANY OPTIONS TO IT
\usepackage{caption} % DO NOT CHANGE THIS AND DO NOT ADD ANY OPTIONS TO IT
\usepackage{algorithm}
\usepackage{algorithmic}

\usepackage{subfigure}
\usepackage{graphicx}
\usepackage{bm}  
\usepackage{amsmath} 
\usepackage{makecell}

\usepackage{caption}
\usepackage{amssymb} 

\usepackage[para]{threeparttable} 
\usepackage{newfloat}
\usepackage{listings}
\DeclareCaptionStyle{ruled}{labelfont=normalfont,labelsep=colon,strut=off} % DO NOT CHANGE THIS
\floatstyle{ruled}
\newfloat{listing}{tb}{lst}{}
\floatname{listing}{Listing}
\nocopyright 
\title{PADiff: Predictive and Adaptive Diffusion Policies for Ad Hoc Teamwork}
\author {
    Hohei Chan\textsuperscript{\rm 1},
    Xinzhi Zhang\equalcontrib \textsuperscript{\rm 1},
    Antao Xiang\textsuperscript{\rm 1},
    Weinan Zhang\textsuperscript{\rm 2},
    Mengchen Zhao\thanks{corresponding author.}\textsuperscript{\rm 1}
}
\affiliations {
    \textsuperscript{\rm 1}School of Software Engineering, South China University of Technology, Guangzhou, China\\
    \textsuperscript{\rm 2}Shanghai Jiao Tong University, Shanghai, China\\
    hoheichanchx@gmail.com, zhang.xinzhi@outlook.com, 202330421721@mail.scut.edu.cn, wnzhang@sjtu.edu.cn, zzmc@scut.edu.cn
}

\begin{document}

\maketitle

\begin{abstract}
Ad hoc teamwork (AHT) requires agents to collaborate with previously unseen teammates, which is crucial for many real-world applications. The core challenge of AHT is to develop an ego agent that can predict and adapt to unknown teammates on the fly. Conventional RL-based approaches optimize a single expected return, which often causes policies to collapse into a single dominant behavior, thus failing to capture the multimodal cooperation patterns inherent in AHT. In this work, we introduce PADiff, a diffusion-based approach that captures agent's multimodal behaviors, unlocking its diverse cooperation modes with teammates. However, standard diffusion models lack the ability to predict and adapt in highly non-stationary AHT scenarios. To address this limitation, we propose a novel diffusion-based policy that integrates critical predictive information about teammates into the denoising process. Extensive experiments across three cooperation environments demonstrate that PADiff outperforms existing AHT methods significantly. Code is available at https://github.com/Chancc235/PADiff.
\end{abstract}

\section{Introduction}

Ad hoc teamwork (AHT) presents a core challenge in multi-agent systems \cite{stone2010ad}, which requires agents to collaborate with unknown teammates without predefined coordination \cite{yuan2023learning}. Consider a robotic soccer match, where an autonomous agent plays with unfamiliar teammates. The agent must make real-time decisions without any knowledge of the new teammates’ playing styles or strategies. These challenges are not limited to simulated environments. In real-world applications like disaster response \cite{yucesoy2025role}, agents must coordinate with human partners operating under unknown procedures. In autonomous driving \cite{teng2023motion}, vehicles often need to interact with other road users without established communication protocols. Such situations require agents to infer teammate behaviors, adapt their strategies, and cooperate effectively on the fly \cite{ravula2019ad}. As multi-agent systems are deployed in real world \cite{yuan2023survey}, it is urgent to address the challenges of AHT and improve the autonomy of multi-agent systems. 

\begin{figure*}[ht]
\vskip 0.2in
\begin{center}
\centerline{\includegraphics[scale=0.66]{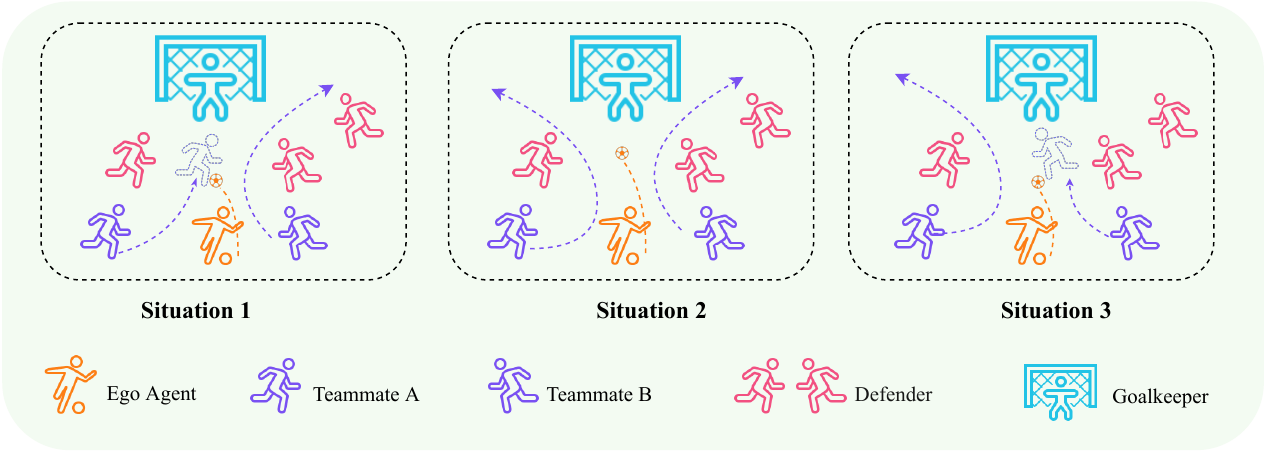}}
\caption{An example illustrating multiple cooperation patterns in AHT. Situation 1: The ego agent passes the ball to teammate A; Situation 2: The ego agent attempts a shot directly; Situation 3: The ego agent  passes the ball to teammate B.}
\label{fig: multi-modal}
\end{center}
\vskip -0.3in
\end{figure*}

AHT fundamentally requires an ego agent to collaborate with previously unseen teammates, making it crucial to prepare for multiple possible cooperative patterns rather than relying on a single optimal response. Consider a robotic soccer scenario involving AHT shown in Figure \ref{fig: multi-modal}): When an ego agent advances with the ball, she has to choose between two modes of cooperation: either assist a teammate or directly attempt a shot. This example indicates that the ego agent must learn multiple cooperation patterns simultaneously, highlighting the multimodal nature of the optimal policies. However, most existing solutions to AHT are based on reinforcement learning (RL) \cite{chen2020aateam, gu2021online,papoudakis2021agent}, which often fail to learn multimodal policies. RL typically optimizes for a fixed expected return, causing the policy to collapse to a single dominant behavior. From the generalization perspective, even if RL learns an optimal policy, it often still fails to generalize when the teammates change their cooperation patterns. Although previous AHT methods (e.g., LIAM\cite{papoudakis2021agent} and GPL-SPI\cite{rahman2021towards}) leverage maximum entropy RL (e.g., SAC) to enhance stochastic exploration and mitigate policy collapse into a singular cooperative mode, their reliance on undirected behavioral dispersion inherently fails to structurally model multimodal distributions, thus poorly capturing diverse cooperation strategies. In contrast, diffusion models \cite{croitoru2023diffusion} are naturally suitable to model multimodal distributions\cite{wang2022diffusion}, making them stronger potential solutions to AHT, where learning multiple cooperation patterns simultaneously is essential for robust teamwork.

Diffusion models, serving as policies, while effective in capturing multimodal action distributions, face two main challenges when applied to AHT. (1) While diffusion models excel at generating diverse actions through iterative denoising, they are primarily designed for distribution reconstruction and lack the predictive ability required for effectively supporting decision-making. This limitation raises challenges for applying diffusion models to AHT, where anticipating teammate intentions and making context-aware decisions are crucial. (2) Traditional diffusion models are typically built on MLP or UNet architectures, which lack the architectural flexibility to adapt to changing teammate behaviors. This presents a limitation for AHT, where real-time adaptation to dynamic and unpredictable teammates is essential for effective collaboration.
%In summary, the main limitation of diffusion models in AHT lies in their lack of predictive and adaptive capabilities, which are essential for effective decision-making and real-time adaptation to changing teammates.

To address these two limitations, we present PADiff, a diffusion-based framework for learning Predictive and Adaptive policies in AHT. The core innovations of PADiff are as follows. (1) To address the lack of predictive capability in traditional diffusion models, we design a \textbf{Predictive Guidance Block (PGB)} integrated into the denoising process. This module  specifically leverages intermediate representations to predict teammates’ cooperative targets and align action generation with long-term team objectives. (2) To address the limited adaptability of traditional diffusion architectures, we design an \textbf{Adaptive Feature Modulation Net (AFM-Net)}, which integrates two FiLM-like feature-wise modulation layers to scale and shift intermediate features based on the current state \cite{perez2018film}. To further improve the diffusion policies' stability and adaptability against uncertain teammates, AFM-Net integrates several designs such as Layer Normalization, Residual connections and Dropout regularization into the training process. Through this way, AFM-Net dynamically adjusts its internal representations to accommodate the frequently changing teammates' goals and behaviors.  

The main contributions of our work can be summarized as follows. 
\begin{enumerate}
    \item We introduce PADiff, which is the first diffusion-based approach to the problem of AHT. PADiff enables the agent to learn multimodal cooperation patterns, leading to diverse collaboration with unknown teammates. This addresses the limitation of traditional RL-based methods which collapse to a single dominant behavior.
    \item PADiff integrates two novel modules to enhance the diffusion policies' adaptability and prediction ability. First, the AFM-Net enables real-time adaptation to non-stationary teammates, ensuring robust performance in dynamic AHT scenarios. Second, the PGB effectively guides the denoising process during training by predicting teammates’ cooperative targets.
    \item We empirically validate PADiff in three classic collaborative environments. Experimental results demonstrate that PADiff consistently outperforms strong baselines in diverse scenarios, achieving an impressive average performance gain of 35.25\% across all tested environments.
\end{enumerate}

\begin{figure*}[ht]
% \vskip 0.2in
\begin{center}
\centerline{\includegraphics[scale=1.2]{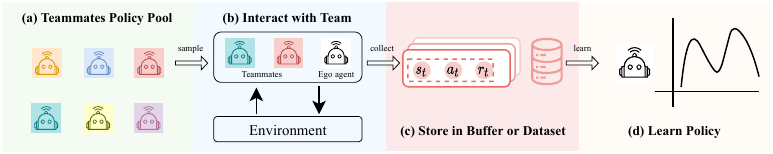}}
\caption{Overview of the AHT training pipeline. The ego agent learns to cooperate with diverse teammates by: (a) sampling from a heterogeneous teammate policy pool to simulate varied collaboration scenarios, (b) interacting within the environment to collect trajectories, (c) storing interaction data, and (d) continuously optimizing its policy based on the collected data.}
\captionsetup{font=small}
\label{fig: pipeline}
\end{center}
\vspace{-2em}
\end{figure*}

\section{Related works}

\textbf{Ad Hoc Teamwork.} The core challenge of AHT is to develop a policy that generalizes across diverse teammate behaviors and enables rapid adaptation during execution. 
Early approaches typically rely on predefined sets of teammate types, using type inference to switch among corresponding policies (e.g., PLASTIC \cite{barrett2015cooperating}). While effective in constrained environments, these methods \cite{chen2020aateam,barrett2015cooperating,durugkar2020balancing,mirsky2020penny} struggle when teammates exhibit complex or unmodeled behaviors. 
More recent works \cite{papoudakis2021agent,gu2021online} improve adaptability by learning representations of teammates, allowing the agent to adjust its policy based on online interactions. These methods relax the assumption of fixed types and broaden the space of potential collaborations. 
However, most existing approaches \cite{barrett2017making, chen2020aateam, rahman2021towards, zhang2025ad} are still built on RL frameworks, where the optimization objective focuses on maximizing expected return. This leads the learned policy to converge toward a single high-reward action, neglecting the diversity of possible cooperative modes under different teammate dynamics. While maximum entropy regularization (\cite{haarnoja2018soft}) mitigates this collapse by promoting exploratory behavior, it fundamentally fails to construct genuinely multimodal policies. 
Such limited policy expressiveness directly constrains generalization in AHT. In contrast, diffusion models \cite{zhu2023diffusion} have the ability to represent any normalizable distribution, with the potential to improve the AHT policy representation effectively.\\

\noindent\textbf{Generative Models for Decision Making.} Generative Models for Decision Making. Generative models have seen increasing use in decision-making tasks \cite{gozalo2023chatgpt,jo2023promise,brynjolfsson2025generative,goodfellow2020generative}.
While GAN-based methods \cite{goodfellow2020generative} like GAIL \cite{ho2016generative} offer adversarial imitation, they often struggle with training instability and limited generalization. VAEs \cite{kingma2013auto,mandlekar2020learning,mees2022matters,lynch2020learning} utilize latent representations for policy generation but face difficulties capturing fine-grained dynamics. More recently, Transformer-based models \cite{chen2021decision,zheng2022online,xie2023future}, notably Decision Transformer \cite{chen2021decision}, have reframed policy learning as return-conditioned sequence modeling, achieving strong offline RL performance.
However, most generative policy models remain unimodal due to architectural biases like Gaussian priors and autoregressive decoding. In contrast, diffusion-based policies \cite{wang2022diffusion} can learn multimodal action distributions. Leveraging this, we embed teammate-predictive guidance into a diffusion-based policy framework specifically for AHT. 
\vspace{6pt}

\noindent\textbf{Diffusion Models.} Diffusion models \cite{sohl2015deep}, first for general high-dimensional data, rapidly found success in image synthesis. Their strength here lies in U-Net backbones' multi-scale feature aggregation and spatial inductive bias
\cite{ho2020denoising,nichol2021improved,vahdat2021score}. DiT \cite{peebles2023scalable} replaces the U-Net backbone with Vision Transformers, showing that attention mechanism improves sample quality by capturing long-range dependencies. Beyond vision, diffusion models have recently been adopted for decision-making in RL field. Trajectory-level planners 
\cite{janner2022planning,liang2023adaptdiffuser,liang2024skilldiffuser,zhang2024motiondiffuse} generate full state–action sequences for long-horizon reasoning, typically with U-Nets \cite{janner2022planning} or Transformer variants like DiT \cite{zhang2024motiondiffuse}. Action-level policies, by contrast, denoise one action at a time with lightweight MLPs \cite{wang2022diffusion,kang2023efficient,hansen2023idql}, enabling real-time control. 
We follow the action-level paradigm but tackle the more demanding AHT setting. To this end, we introduce (i) a \emph{Conditional Feature Modulation Network} that sustains robustness under dynamically changing teammate policies and (ii) a \emph{Predictive Guidance Block} that injects teammate-aware goals into the denoising process.

\section{Preliminaries}
\textbf{Ad Hoc Teamwork.}
We consider training an ego agent to collaborate with unknown teammates in cooperative multi-agent environments to achieve a shared goal. This problem is modeled as a Decentralized Markov Decision Process (Dec-MDP) extended with a teammate policy space: 
$\langle N, \mathcal{S}, \mathcal{A}, P, R, \Gamma, \mathcal{T} \rangle,$
where $N = \{1, 2, \dots, n\}$ denotes the set of agents, $\mathcal{S}$ is the global state space, and the joint action space is defined as $\mathcal{A} = \mathcal{A}^1 \times \dots \times \mathcal{A}^N$. Without loss of generality, we designate agent $i$ as the ego agent and $-i$ as the set of teammates. The transition function $P: \mathcal{S} \times \mathcal{A} \rightarrow \Delta(\mathcal{S})$ governs the environment dynamics, and $R$ is the shared reward function. $\Gamma$ represents the joint policy space of all potential teammates. We define a trajectory $\tau = \{(\mathbf{s}_t, \mathbf{a}_t, r_t)\}_{t=0}^{T}$ and denote by $\mathcal{T}$ the set of such episodic trajectories generated from interactions between the ego agent and sampled teammates.

As illustrated in Figure~\ref{fig: pipeline}, the AHT training pipeline contains four stages:
(a) Sample a subset of policies from a diverse teammate pool at the beginning of each episode;
(b) Interact with the ego agent to collect full trajectories $\tau$;
(c) Store trajectories in the offline dataset;
(d) Optimize the ego policy $\pi_{\theta}^i(a^i \mid s; \mathcal{D})$ based on the collected data $\mathcal{D}$.\\

\noindent\textbf{Diffusion Probabilistic Models.}
Diffusion models (DMs) \cite{sohl2015deep,song2019generative} are a powerful type of generative models that learn to recover data samples from noise. This typically involves two phases: a forward noising process that gradually perturbs clean data into noise, and a reverse process that learns to reconstruct data from noise.
Given a data sample \(x_0 \sim p_{\text{data}}(x)\), the forward process defines a Markov chain:
$p(x_k \mid x_{k-1}) = \mathcal{N}(x_k \mid \sqrt{\alpha_k} x_{k-1}, (1 - \alpha_k) I),$ where \(\alpha_k\) controls the noise level at step \(k\). The reverse process is parameterized by:
$q_\theta(x_{k-1} \mid x_k) = \mathcal{N}(x_{k-1} \mid \mu_\theta(x_k, k), (1 - \alpha_k) I).$ Sampling starts from \(x_K \sim \mathcal{N}(0, I)\) and iteratively denoises toward \(x_0\). The model is trained by minimizing the loss:
\begin{equation}
\mathcal{L}(\theta) = \mathbb{E}_{k, x_0, \epsilon} \left[ \| \epsilon - \epsilon_\theta(x_k, k) \|^2 \right], \quad \epsilon \sim \mathcal{N}(0, I).
\label{eq:ddpm-loss}
\end{equation}

\begin{figure*}[!ht]
\begin{center}
% \vspace{-10mm}
% \centerline{\includegraphics[scale=0.9]{Method6.pdf}}
\includegraphics[width=0.85\textwidth]{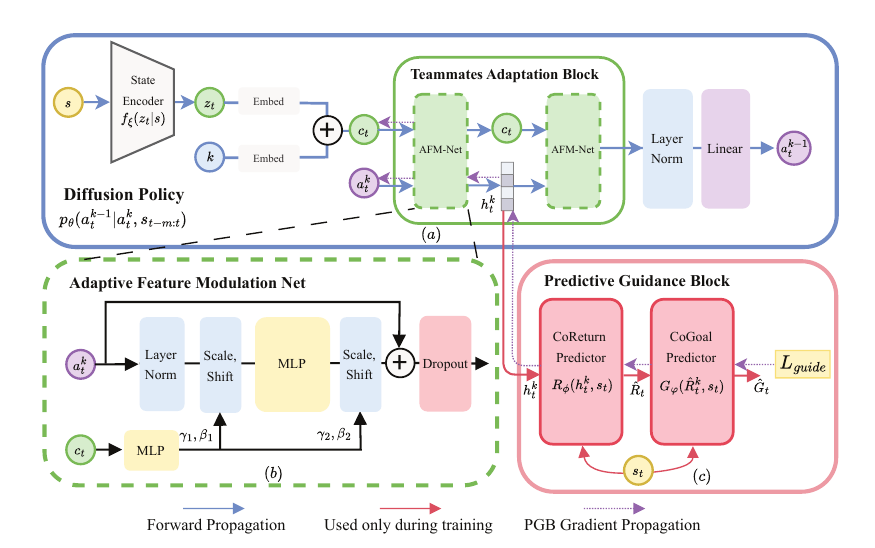}
    \captionsetup{font=small}
    \caption{The overall architecture of PADiff.  
    The subgraph (a) shows that PADiff represents the policy as a diffusion-based model and the State Encoder $f_\xi(z_t | s_{t-m:t})$ transforms states into latent representation capturing team dynamics. $c_t$ represents the vector obtained by bit-wise addition of the embedding of latent representations $z_t$ and the embedding of diffusion step $k$, which serves as the condition for diffusion.
    The subgraph (b) showcases our AFM-Net uses $c_t$ representing teamwork context  as the denoising condition to dynamically modulate the intermediate feature vectors to affect the actions generation.
    The subgraph (c) illustrates that PGB integrates the team-aware information into the action denoising process by predicting teammates' intentions while training, through gradient propagation, ensuring that the ego agent can make such decisions that align with long-term team objectives while testing.
    } 
    \label{fig:method}  
\end{center}
\vspace{-6mm}
\end{figure*}

\section{Methods}

\textbf{Overview.}
Our framework PADiff leverages a diffusion-based approach tailored for cooperative decision-making in AHT environments. As illustrated in Figure \ref{fig:method}, PADiff comprises three key components: (1) diffusion-based policy representation, (2) Teammates Adaptation Block, and (3) Predictive Guidance Block (PGB). Initially, the state encoder transforms the global state into a latent variable encoding the current teamwork context. The latent representation, together with action embedding, enters the Teammates Adaptation Block, consisting of multiple Adaptive Feature Modulation Nets (AFM-Net), which dynamically adjusts feature vectors via a FiLM-based conditioning mechanism, generating adaptive actions responsive to teammate dynamics. Simultaneously, the PGB predicts collaborative return and collaborative goal from intermediate features $h_t^k$ of Teammates Adaptation Block, guiding the denoising process toward optimal team-aligned objectives by integrating predictive information via gradient signals.

During training, these three components jointly enable the policy model to have both predictive and adaptive ability. At inference, since the denoising module internalizes team-awareness, the PGB is no longer required, enabling efficient real-time adaptive decision-making. The procedure can be found in Alg1 and Alg2 of the Appendix. In the following, we detail each core component of PADiff.

\subsection{Overall Diffusion Policy Representations}
\begin{figure}[h!]
    \vspace{-5pt}
    \centering
    \includegraphics[width=0.4\textwidth]{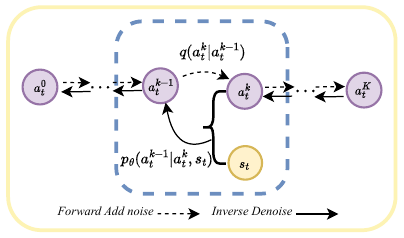}
    \caption{Forward adding noise process and inverse denoising process in PADiff.
    }
    \label{fig:diffusion_process}
    \vspace{-8pt}
\end{figure}

To capture agent's potential multimodal behaviors required for effective ad hoc teamwork, unlocking its diverse cooperation modes with teammates, we model the ego agent’s policy as a conditional diffusion process. The iterative denoising framework naturally accommodates complex conditions, enabling us to inject compact teammates information at each step. We define the policy distribution over actions $\bm{a}_t$ given state $s_t$ as the reverse diffusion denoising chain:
\begin{equation}
\pi_\theta(\bm{a}_t|s_t) = p_\theta(\bm{a}_t^{0:K}|s) = \mathcal{N}(\bm{a}_t^K;0,\bm{I})\prod_{k=K}^0 p_\theta(\bm{a}_t^{k-1}|\bm{a}_t^k,s),
\end{equation}
As shown in figure~\ref{fig:diffusion_process}, the solid lines show an inverse diffusion process where a noisy action $\bm{a}_t^K$ sampled from a Gaussian distribution progressively transforms into the final executable action $\bm{a}_t^0$ through K diffusion steps while execution.

For discrete action spaces, following D3PM \cite{austin2021structured}, the reverse diffusion is parameterized as:
\begin{equation}
\label{eq:denoise_dis}
\bm{a}^{k-1} | \bm{a}^k \sim p_\theta(\bm{a}^{k-1}|s, \bm{a}^k), \quad k = K, \dots, 1,
\end{equation}
and the forward diffusion process can be represented as:
\begin{equation}
   q(\bm{a}_t^{k}|\bm{a}_t^0) = \text{Cat}(\bm{a}_t^k;p=\bm{a}_t^0\overline{Q}k).
\label{eq:add_noise1}
\end{equation}
\begin{equation}
   q(\bm{a}_t^k|\bm{a}_t^{k-1}) = \text{Cat}(\bm{a}_t^k;p=\bm{a}_t^{k-1}Q_k)
\label{eq:add_noise2}
\end{equation}
where $\overline{Q}_k=Q_1\cdot Q_2\dots Q_k$ and the schedule type of transition matrices $Q$ follows the Uniform approach, which is a way of controlling the forward diffusion process and the learnable reverse diffusion process. $\text{Cat}(\bm{x};\bm{p})$ is a categorical distribution over the one-hot row vector $\bm{x}$ with probabilities given by the row vector $\bm{p}$ \cite{austin2021structured}.

During training, we sample $k$ from $\{1,\dots, K\}$. As shown in figure~\ref{fig:diffusion_process}, the dotted lines direction shows that the sample is prepared by starting with ground truth action $\bm{a}_t^0$. We can compute $\bm{a}^k$ and $\bm{a}^{k-1}$ using equation \ref{eq:add_noise2}.  Then the model learns to model the entire distribution $p_\theta(\bm{a}_t^{k-1} | \bm{a}_t^k, s_t)$ by optimizing the variational lower bound. The loss function for training is as follows:
\begin{equation}
\label{eq:diff_loss2}
    \begin{split}
    \mathcal{L}_{Diff} =\mathbb{E}\left[  
    D_{KL}\left[ q(\bm{a}^{k-1}|\bm{a}^k,\bm{a}^0 ) \|p_{\theta}(\bm{a}^{k-1}|\bm{a}^{k},s,k)\right] \right]\\
    -  \log p_\theta(\bm{a}^0|\bm{a}^1,s) 
    \end{split}
\end{equation}
where \begin{flalign*}
    q(\bm{a}_t^{k-1}|\bm{a}_t^k,\bm{a}_t^0) = \frac{q(\bm{a}_t^{k}|\bm{a}_t^{k-1},\bm{a}_t^0 )q(\bm{a}_t^{k-1}|\bm{a}_t^0 )}{q(\bm{a}_t^{k}|\bm{a}_t^0 )} &\\
    = \text{Cat}\left(\bm{a}_t^{k-1};\bm{p}=\frac{{\bm{a}_t^k}{Q^\top _k} \,\odot \, \bm{a}_t^0 \overline{Q}_{k-1}}{\bm{a}_t^0 \overline{Q}_{k}{\bm{a}_t^k}^\top}\right).&
\end{flalign*}

While our diffusion policy captures the full action distribution, standard diffusion models are insufficient for sophisticated ad hoc teamwork. They typically lack (1) efficient adaptation to dynamic teammates and (2) integrated teammate intent prediction. PADiff addresses these limitations with its novel Adaptive Feature Modulation Net and Predictive Guidance Block, which we will detail in the following.

\subsection{State-Conditioned Teammates Adaptation}
\label{sec:adaptive}
The denoising process $p_\theta(\bm{a}_t^k,s_t,k)$ uses a state encoder and the Teammates Adaptation Block.
The state encoder compresses the state into a more informative latent variable. Then the latent variable, diffusion step and the action embedding are fed into the Teammates Adaptation Block, which is consisting of two AFM-Net, simultaneously to generate the denoised action. 

\textbf{State Encoder:}
To better identify the cooperation situation of the team and handle potential changes in uncertain teammates, we model the team's cooperation state using a history window $s_{t-m:t}$ which provides richer context about interactions. This sequence is compressed into an informative latent space modeled as a multivariate Gaussian distribution \cite{gu2021online}. A sample from this latent distribution acts as the conditional input during the denoising stage.
\begin{equation}
    (\mu_{z_t},\sigma_{z_t}) = f_\xi(s_{t-m:t}),\, z_t \sim \mathcal{N}(\mu_{z_t},\sigma_{z_t})
\end{equation}

\textbf{Adaptive Feature Modulation Net (AFM-Net):}
Traditional denoising networks, such as UNet and MLP, lack necessary teammate-awareness, while Transformer-based networks like DiT\cite{peebles2023scalable} rely on attention mechanisms that introduce significant computational overhead, undesirable in fast-paced AHT settings. To overcome these limitations, we propose AFM-Net, an effective and robust denoising network that integrates the FiLM mechanism, residual connections, layer normalization and dropout regularization, making it both expressive and efficient without the need for attention.
AFM-Net has three primary features: (1) Conditional feature modulation through FiLM-style\cite{perez2018film} method, enhancing teammate adaptability, (2) Residual connections for stable training and robust representations and (3) Dropout regularization to improve generalization to unseen teammates. 

Specifically, we generate $\gamma_j$(scale) and $\beta_j$(shift) parameters from the team context $z_t$ using an MLP. These parameters dynamically modulate the intermediate feature vectors produced by Layer Norm Block and MLP Block. This integration enables the model to generate actions that adapt to the team. Additionally, residual connections and dropout regularization ensure stable learning and enhance the model's ability to generalize to new, unseen teammates.

\begin{equation}
    \gamma_1, \beta_1, \gamma_2, \beta_2 = MLP(\bm{z}_t + k)
\end{equation}
\begin{equation}
\begin{split}
    \textbf{AFM}(\bm{x}, z_t, k) = \gamma_2\cdot(MLP(\gamma_1\cdot LN(\bm{x})+\beta_1)) \\
    + \beta_2 + \bm{x}
\end{split}
\end{equation}

where $\cdot$ represents the element-wise dot product, $\bm{x}$ represents the intermediate feature vector, LN represents the Layer Normalization module.

\begin{figure*}[ht]
\begin{center}
\vspace{-10pt}
\centerline{\includegraphics[scale=0.05]{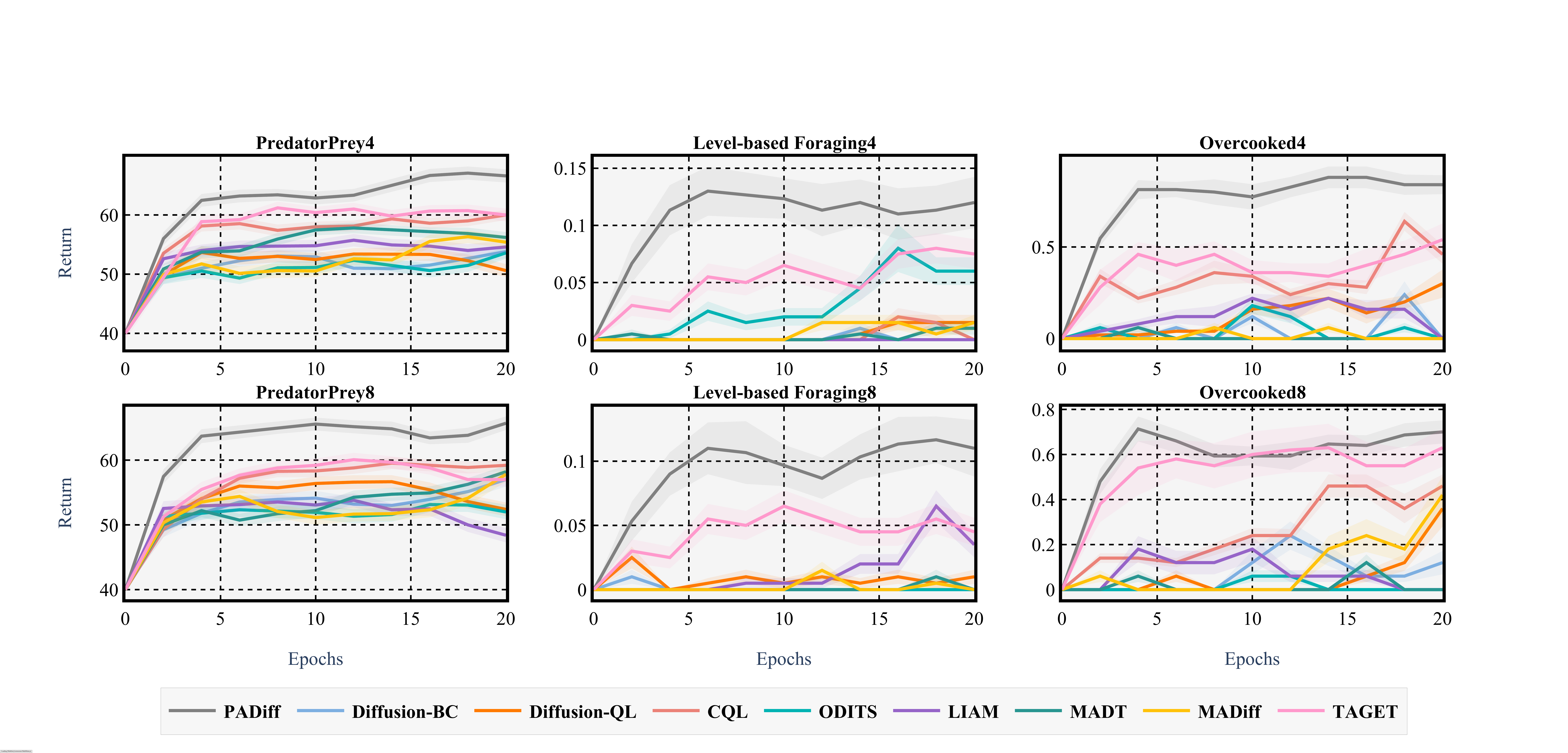}}
    \caption{Average evaluation returns with 95\% confidence interval during training.
    } 
    \label{fig:main}  
\end{center}
\vspace{-10mm}
\end{figure*}

\subsection{Predictive Denoising with Collaborative Goal Guidance}
\label{sec:predictive}
To address the challenge that existing diffusion-based policies lack the ability to predict teammates’ intentions and struggle to adapt to the behaviors of previously unseen teammates, we propose the \textbf{Predictive Guidance Block (PGB)}, a novel component integrated into the denoising phase of the PADiff framework, which can integrate predictive information into the denoising 
process. 

Unlike conventional inference-time guidance methods, which often lack flexibility in adapting to real-time teammate dynamics, PGB leverages the intermediate representation generated during the denoising process to predict teammates' intentions. By propagating gradients through the denoising process, it optimizes this representation, enabling it to anticipate teammates' behaviors. As a result, the denoising process becomes adaptive to changes in teammates' actions. 
The design of the PGB contains two core predictive tasks: \textbf{Collaborative Return (CoReturn)} represents the expected cumulative future team reward, conditioned on the current state and intermediate feature from the denoising network. The prediction of CoReturn serves to guide the generation of actions towards better alignment with overall team objectives. And \textbf{Collaborative Goal (CoGoal)} stands for the predicted future state, which is served as a sub-goal to capture the intention of teammates. 
While model-base method like DreamerV3\cite{hafner2023mastering} optimize policies through environment prediction, PGB similarly enhances decision-making via auxiliary prediction tasks, but fundamentally differs by tailoring predictions to real-time collaborative intent. Each predictive task contributes distinctively to enhanced robust cooperative decision-making.

During the training phase, we predict CoReturn $R_t$ using the meaningful intermediate features $\bm{h}_t^k$ directly produced by AFM-Net and state $s_t$, and then use CoReturn along with $s_t$ to predict CoGoal $G_t$.
This sequential prediction process is designed to model the hierarchical nature of team coordination. The initial prediction of CoReturn provides an evaluation of the expected reward, which serves as a foundational signal for determining the next sub-goals, which are captured by CoGoal. The loss function is defined as:

\begin{equation}
\label{eq:rtg}
L_{\text{CoReturn}} = \mathbb{E}_{\tau \sim \mathcal{D}} \left[ \sum_{t=1}^{T} \| R_{\phi}(h^k_t, s_t) - R_t \|^2 \right],
\end{equation}

\begin{equation}
\label{eq:sub-goal}
\begin{split}
    L_{\text{CoGoal}} = \mathbb{E}_{\tau \sim \mathcal{D}} \Bigg[ - \frac{1}{N}  \sum_{t=1}^T \sum_{i=1}^N \left( G_{t,i} \log \hat{G}_{t,i} \right. \\
+ \left. (1 - G_{t,i}) \log (1 - \hat{G}_{t,i}) \right) \Bigg]
\end{split}
\end{equation}
\noindent where $\hat{G}_t = G_{\varphi}(\hat{R}^k_t, s_t),\, G_{t} = s_{t+n}, \,R_t=\sum_{t'=t}^{T}r_{t'}$. We perform one-hot encoding on future state $s_{t+n}$,  where each dimension represents a binary indicator. We use BCE loss per dimension to handle binary classification, which aligns with the discrete nature of our environments.

The total loss is represented as:
\begin{equation}
\label{eq:total_loss}
    L_{\text{total}} = L_{\text{Diffusion}} + \alpha L_{\text{CoReturn}} + \beta L_{\text{CoGoal}}
\end{equation}

Formally, let $\nabla_{h_t^k} L_{\text{CoReturn}}$ and $\nabla_{h_t^k} L_{\text{CoGoal}}$ represent the gradient derived from each prediction objective. The total gradient update on $h_t^k$ can be represented as:
\begin{equation}
\nabla_{h_t^k}L_{\text{total}} = \nabla_{h_t^k}L_{\text{Diffusion}} + \alpha \nabla_{h_t^k}L_{\text{CoReturn}} + \beta \nabla_{h_t^k}L_{\text{CoGoal}},
\end{equation}
where $\alpha, \beta$ are weighting hyperparameters. By minimizing prediction errors on CoReturn and CoGoal, intermediate features become predictive of beneficial cooperative targets, ensuring that the ego agent can respond effectively to teammates' behaviors. During the inference phase, the Predictive Guidance Block is no longer required, as the denoise block AFM-Net has learned the capacity to produce intermediate features that are predictive of team-aware behavior, which enables the denoising block to embed team-awareness into its action generation and maintains effective coordination capacity without real-time guidance.

\section{Experiments}
\label{sec:exp}
\textbf{Setup.} To better evaluate the performance of the proposed model in Ad Hoc Teamwork scenarios, we employ three benchmark environments: \textbf{Predator-Prey (PP)}, \textbf{Level-Based Foraging (LBF)}, and \textbf{Overcooked}. Predator-Prey: In a grid-world environment, a team of predator agents must collaborate to capture a prey that actively attempts to evade them. Successful captures require tight coordination among predators.
Level-Based Foraging: Agents operate in a grid world where they collect food items randomly distributed across the environment. Both agents and food items are assigned discrete levels, and a food item can only be collected when the combined levels of participating agents meet or exceed the food's level. 
Overcooked: Agents must efficiently prepare and deliver dishes in a kitchen with a complex spatial layout. Effective cooperation with teammates and synchronized movements are crucial. 

\textbf{Collections of teammates policies.} In AHT scenarios, we must ensure that the ego agent can adapt to previously unseen teammates. To build a diverse pool of teammate policies, we use the Soft-Value Diversity (SVD) method from the CSP\cite{ding2023coordination} framework to collect teammate behaviors, instantiating four independent multi‑agent populations in each environments. The training process alternates between the inner loop and the outer loop. In the inner loop, each team is trained in isolation, updating its policy to maximize its own performance. In the outer loop, we jointly update all teams to maximize the SVD objective, encouraging them to develop distinct value estimates over observation–action pairs, thus diversifying their behaviors. Specifically, three training populations are dedicated to interacting with the ego agent for collecting the training data, while a fourth population is reserved exclusively for assessing the ego agent’s performance. Each training population comprises three unique policy checkpoints, whereas the testing population includes twelve checkpoints. These checkpoints correspond to distinct joint policies. By isolating the training and testing in this way, our protocol ensures a rigorous evaluation of the model’s ability to adapt to previously unseen teammate strategies. We divided the 12 test policies into two groups, one containing 4 test strategies and the other containing 8. The cross-play matrices experiments reveal that cooperation scores for different teammate combinations are lower than for identical teammate combinations. This is due to strategy differences causing coordination issues, which reflects the diversity of test teammates strategies. Detailed information about cross-play experiments can be found in Appendix A.2. 

\textbf{Baselines.} In our experiments, we compare two classic AHT methods: \textbf{LIAM} \cite{papoudakis2021agent} and \textbf{ODITS} \cite{gu2021online}. To underscore the benefits of our method and its superior adaptability to AHT tasks, we further benchmark it against two diffusion‑based models: \textbf{Diffusion‑BC} and \textbf{Diffusion‑QL} \cite{wang2022diffusion}. We also compare with the offline reinforcement learning algorithm \textbf{CQL} \cite{kumar2020conservative} and two AHT–focused multi‑agent methods: \textbf{MADIFF} \cite{zhu2024madiff}, which aggregates all agents’ observations into a global state sequence and uses an inverse dynamics model to predict the ego agent’s next action; and \textbf{MADT} \cite{meng2021offline}, which only trains the ego agent’s actor network within the AHT settings. We also take the latest DT-based offline AHT method \textbf{TAGET}  \cite{zhang2025ad} as baseline. These comparisons demonstrate that our method addresses the challenges in AHT scenarios more effectively and achieve significant performance.

\subsection{Comparison with Baselines.} We trained our model for 20 epochs in the offline datasets we collected. We evaluate the agent against a held-out test pool every two epochs,. Test pools consist of either 4 or 8 policies. For each environment, one group of policies is drawn at random and paired with the ego agent for evaluation. Returns are averaged over 50 trials, with shaded regions representing 95\% confidence intervals using the standard normal distribution formula: $\bar{x} = \pm 1.96 \cdot \frac{\sigma}{\sqrt{n}}$, where $\sigma$ is the sample standard deviation and $n=50$. In every environment and for both pool sizes, our model consistently surpasses the baseline(cf. Fig \ref{fig:main} and Tab1 of Appendix). Additionally, hyperparameters analysis and detailed network architecture can be found in Fig4 and Fig5 of the Appendix.

\subsection{Visualization of Multimodal Policy Distributions}
To verify that PADiff captures multimodal behaviors and discovers different modes of cooperation, we set up a game scenario in the PP environment and input the same state into the policy multiple times. We found that the ego agent would exhibit different cooperative paths as shown in Fig1 of Appendix, proving that our policy actually fits a multimodal distribution, enabling diverse behaviors under the same state. Further analysis can be found in Appendix A.3.

% \subsection{Comparison among Different Denoising Networks}

\subsection{Ablation Studies}
\begin{figure}[ht]
\begin{center}
\centerline{\includegraphics[scale=0.031]{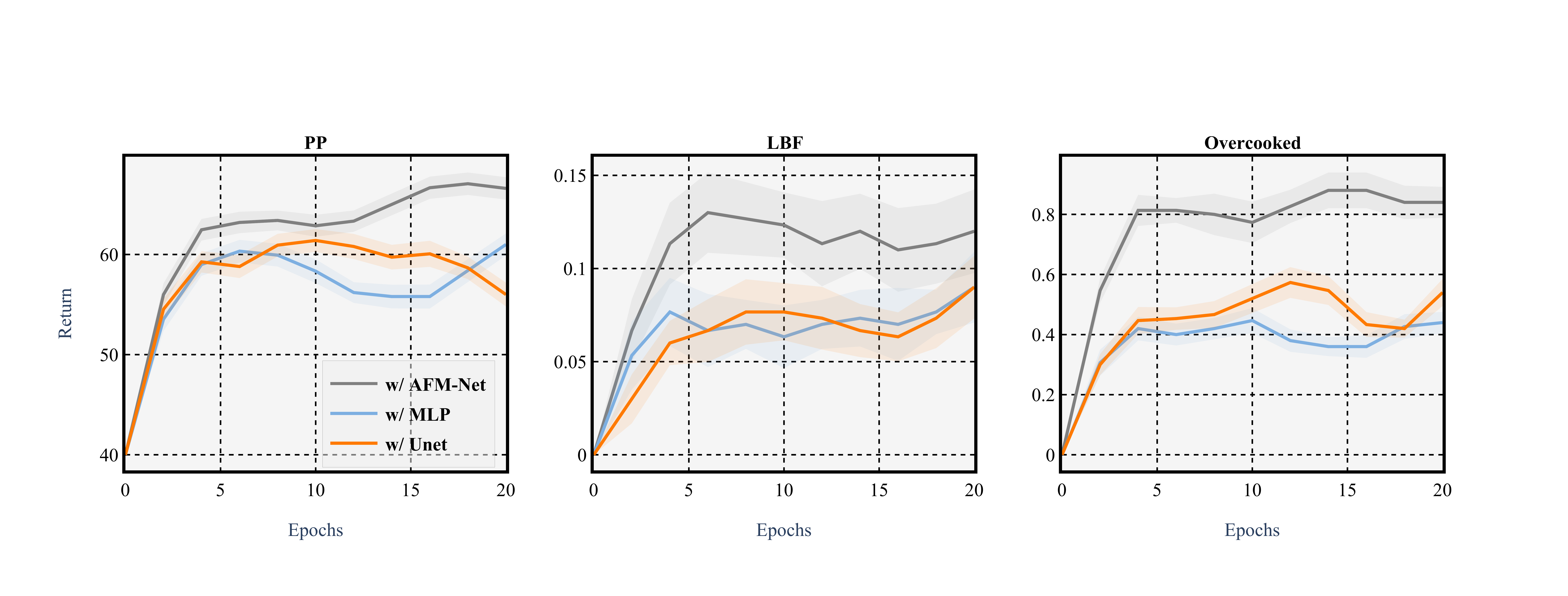}}
    \caption{Ablation results of different denoising networks.} 
    \label{fig:comparison}  
\end{center}
% \vspace{-10pt}
\end{figure}

To validate the necessity of our proposed AFM-Net, we replace AFM-Net in our framework with two alternative architectures: MLP and U-Net, and evaluate performance across all environments.
As shown in the Figure \ref{fig:comparison}, our model consistently outperforms both variants in all environments. These results strongly demonstrate that, in the context of AHT tasks, the design of AFM-Net offers significant advantages and greater applicability compared to a simple MLP and the image-oriented U‑Net architecture.

\begin{figure}[ht]
\begin{center}
\centerline{\includegraphics[scale=0.031]{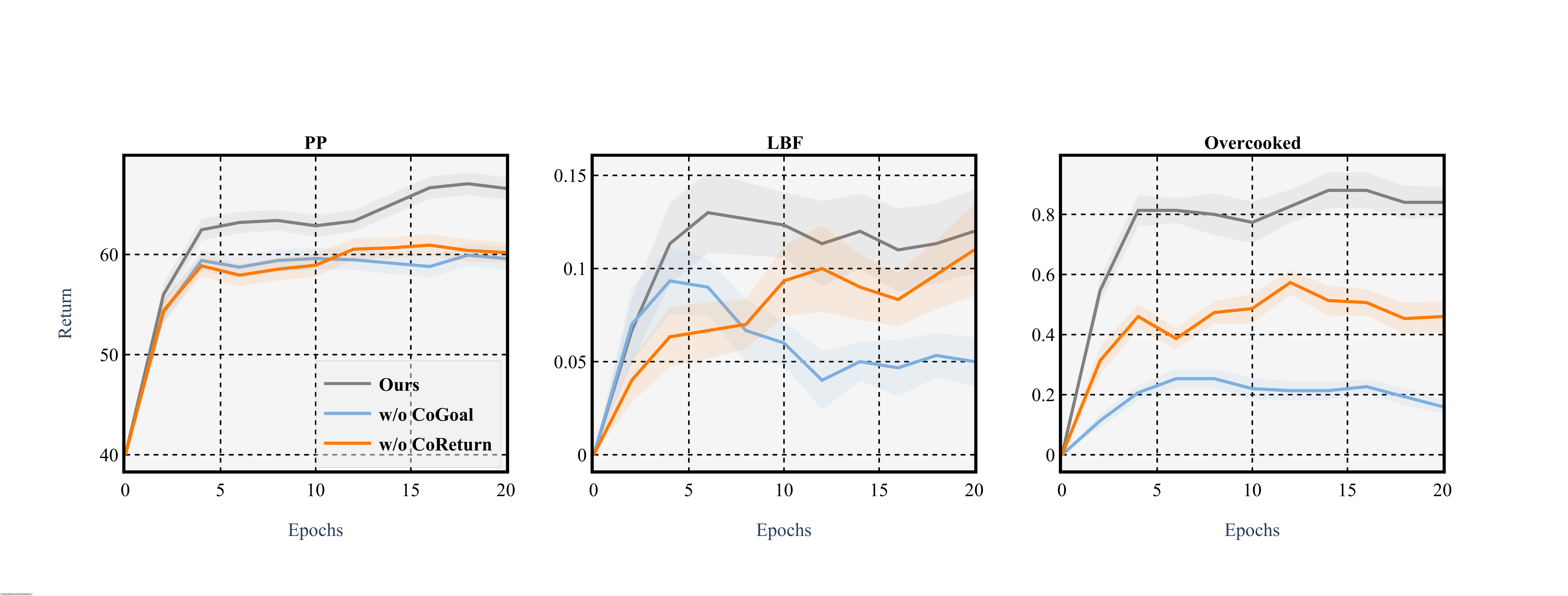}}
    \caption{The ablation results of PGB module.} 
    \label{fig:ablation}  
\end{center}
% \vspace{-10pt}
\end{figure}

To evaluate the contribution of each component of PGB module to overall framework performance, we conducted an ablation study by systematically removing individual modules and assessing performance, allowing us to quantify the impact of each component.
As depicted in the Fig~\ref{fig:ablation}, the results show that the full model consistently outperforms all ablated variants. Both removing the CoGoal predictor and the CoReturn predictor results in substantial degradation, indicating that the ability to predict future team rewards and subgoal are crucial for enabling the ego agent to collaborate with unseen teammates.
These findings strongly support the rationality for the design of PGB, demonstrating that each component plays a vital role in enhancing performance.

\section{Conclusions}
In this work, we introduced PADiff, a novel diffusion-based framework that employs diffusion models as the ego agent’s policy for AHT. Through extensive experiments, we validated that PADiff successfully enhances collaboration in dynamic, unseen teammate scenarios by integrating predictive information into the denoising
process and incorporating adaptive mechanisms for dynamic teammate adaptation. This paves the way for more autonomous and reliable multi-agent systems capable of seamless collaboration in open, unpredictable real-world environments.

\section{Acknowledgments}
This work is supported by National Natural Science Foundation of China (NSFC) under the Youth Science Fund Project (Grant Number: 62506133) and the Guangdong Basic and Applied Basic Research Foundation (Grant Number: 2025A1515010247).

\bibliography{main}

@inproceedings{stone2010ad,
  title={Ad hoc autonomous agent teams: Collaboration without pre-coordination},
  author={Stone, Peter and Kaminka, Gal and Kraus, Sarit and Rosenschein, Jeffrey},
  booktitle={Proceedings of the AAAI Conference on Artificial Intelligence},
  volume={24},
  number={1},
  pages={1504--1509},
  year={2010}
}

@article{teng2023motion,
  title={Motion planning for autonomous driving: The state of the art and future perspectives},
  author={Teng, Siyu and Hu, Xuemin and Deng, Peng and Li, Bai and Li, Yuchen and Ai, Yunfeng and Yang, Dongsheng and Li, Lingxi and Xuanyuan, Zhe and Zhu, Fenghua and others},
  journal={IEEE Transactions on Intelligent Vehicles},
  volume={8},
  number={6},
  pages={3692--3711},
  year={2023},
  publisher={IEEE}
}

@article{barrett2017making,
  title={Making friends on the fly: Cooperating with new teammates},
  author={Barrett, Samuel and Rosenfeld, Avi and Kraus, Sarit and Stone, Peter},
  journal={Artificial Intelligence},
  volume={242},
  pages={132--171},
  year={2017},
  publisher={Elsevier}
}

@article{yuan2023survey,
  title={A survey of progress on cooperative multi-agent reinforcement learning in open environment},
  author={Yuan, Lei and Zhang, Ziqian and Li, Lihe and Guan, Cong and Yu, Yang},
  journal={arXiv preprint arXiv:2312.01058},
  year={2023}
}

@inproceedings{barrett2015cooperating,
  title={Cooperating with unknown teammates in complex domains: A robot soccer case study of ad hoc teamwork},
  author={Barrett, Samuel and Stone, Peter},
  booktitle={Proceedings of the AAAI Conference on Artificial Intelligence},
  volume={29},
  number={1},
  year={2015}
}

@inproceedings{durugkar2020balancing,
  title={Balancing individual preferences and shared objectives in multiagent reinforcement learning},
  author={Durugkar, Ishan and Liebman, Elad and Stone, Peter},
  year={2020},
  organization={International Joint Conference on Artificial Intelligence}
}

@inproceedings{mirsky2020penny,
  title={A penny for your thoughts: The value of communication in ad hoc teamwork},
  author={Mirsky, Reuth and Macke, William and Wang, Andy and Yedidsion, Harel and Stone, Peter},
  year={2020},
  organization={International Joint Conference on Artificial Intelligence}
}

@article{chen2021decision,
  title={Decision transformer: Reinforcement learning via sequence modeling},
  author={Chen, Lili and Lu, Kevin and Rajeswaran, Aravind and Lee, Kimin and Grover, Aditya and Laskin, Misha and Abbeel, Pieter and Srinivas, Aravind and Mordatch, Igor},
  journal={Advances in neural information processing systems},
  volume={34},
  pages={15084--15097},
  year={2021}
}

@article{meng2021offline,
  title={Offline pre-trained multi-agent decision transformer: One big sequence model tackles all smac tasks},
  author={Meng, Linghui and Wen, Muning and Yang, Yaodong and Le, Chenyang and Li, Xiyun and Zhang, Weinan and Wen, Ying and Zhang, Haifeng and Wang, Jun and Xu, Bo},
  journal={arXiv preprint arXiv:2112.02845},
  year={2021}
}

@inproceedings{rahman2021towards,
  title={Towards open ad hoc teamwork using graph-based policy learning},
  author={Rahman, Muhammad A and Hopner, Niklas and Christianos, Filippos and Albrecht, Stefano V},
  booktitle={International conference on machine learning},
  pages={8776--8786},
  year={2021},
  organization={PMLR}
}

@inproceedings{chen2020aateam,
  title={Aateam: Achieving the ad hoc teamwork by employing the attention mechanism},
  author={Chen, Shuo and Andrejczuk, Ewa and Cao, Zhiguang and Zhang, Jie},
  booktitle={Proceedings of the AAAI conference on artificial intelligence},
  volume={34},
  number={05},
  pages={7095--7102},
  year={2020}
}

@phdthesis{ravula2019ad,
  title={Ad-hoc teamwork with behavior-switching agents},
  author={Ravula, Manish Chandra Reddy},
  year={2019}
}

@inproceedings{gu2021online,
  title={Online ad hoc teamwork under partial observability},
  author={Gu, Pengjie and Zhao, Mengchen and Hao, Jianye and An, Bo},
  booktitle={International conference on learning representations},
  year={2021}
}

@article{papoudakis2021agent,
  title={Agent modelling under partial observability for deep reinforcement learning},
  author={Papoudakis, Georgios and Christianos, Filippos and Albrecht, Stefano},
  journal={Advances in Neural Information Processing Systems},
  volume={34},
  pages={19210--19222},
  year={2021}
}

@inproceedings{yuan2023learning,
  title={Learning to Coordinate with Anyone},
  author={Yuan, Lei and Li, Lihe and Zhang, Ziqian and Chen, Feng and Zhang, Tianyi and Guan, Cong and Yu, Yang and Zhou, Zhi-Hua},
  booktitle={Proceedings of the Fifth International Conference on Distributed Artificial Intelligence},
  pages={1--9},
  year={2023}
}

@inproceedings{zheng2022online,
  title={Online decision transformer},
  author={Zheng, Qinqing and Zhang, Amy and Grover, Aditya},
  booktitle={international conference on machine learning},
  pages={27042--27059},
  year={2022},
  organization={PMLR}
}

@article{ding2023coordination,
  title={Coordination Scheme Probing for Generalizable Multi-Agent Reinforcement Learning},
  author={Ding, Hao and Jia, Chengxing and Guan, Cong and Chen, Feng and Yuan, Lei and Zhang, Zongzhang and Yu, Yang},
  year={2023}
}

@inproceedings{xie2023future,
  title={Future-conditioned unsupervised pretraining for decision transformer},
  author={Xie, Zhihui and Lin, Zichuan and Ye, Deheng and Fu, Qiang and Wei, Yang and Li, Shuai},
  booktitle={International Conference on Machine Learning},
  pages={38187--38203},
  year={2023},
  organization={PMLR}
}

@article{kumar2020conservative,
  title={Conservative q-learning for offline reinforcement learning},
  author={Kumar, Aviral and Zhou, Aurick and Tucker, George and Levine, Sergey},
  journal={Advances in Neural Information Processing Systems},
  volume={33},
  pages={1179--1191},
  year={2020}
}

@article{croitoru2023diffusion,
  title={Diffusion models in vision: A survey},
  author={Croitoru, Florinel-Alin and Hondru, Vlad and Ionescu, Radu Tudor and Shah, Mubarak},
  journal={IEEE Transactions on Pattern Analysis and Machine Intelligence},
  volume={45},
  number={9},
  pages={10850--10869},
  year={2023},
  publisher={IEEE}
}

@article{zhu2023diffusion,
  title={Diffusion models for reinforcement learning: A survey},
  author={Zhu, Zhengbang and Zhao, Hanye and He, Haoran and Zhong, Yichao and Zhang, Shenyu and Guo, Haoquan and Chen, Tingting and Zhang, Weinan},
  journal={arXiv preprint arXiv:2311.01223},
  year={2023}
}

@article{wang2022diffusion,
  title={Diffusion policies as an expressive policy class for offline reinforcement learning},
  author={Wang, Zhendong and Hunt, Jonathan J and Zhou, Mingyuan},
  journal={arXiv preprint arXiv:2208.06193},
  year={2022}
}

@article{janner2022planning,
  title={Planning with diffusion for flexible behavior synthesis},
  author={Janner, Michael and Du, Yilun and Tenenbaum, Joshua B and Levine, Sergey},
  journal={arXiv preprint arXiv:2205.09991},
  year={2022}
}

@inproceedings{peebles2023scalable,
  title={Scalable diffusion models with transformers},
  author={Peebles, William and Xie, Saining},
  booktitle={Proceedings of the IEEE/CVF international conference on computer vision},
  pages={4195--4205},
  year={2023}
}

@article{ho2020denoising,
  title={Denoising diffusion probabilistic models},
  author={Ho, Jonathan and Jain, Ajay and Abbeel, Pieter},
  journal={Advances in neural information processing systems},
  volume={33},
  pages={6840--6851},
  year={2020}
}

@article{austin2021structured,
  title={Structured denoising diffusion models in discrete state-spaces},
  author={Austin, Jacob and Johnson, Daniel D and Ho, Jonathan and Tarlow, Daniel and Van Den Berg, Rianne},
  journal={Advances in neural information processing systems},
  volume={34},
  pages={17981--17993},
  year={2021}
}

@inproceedings{perez2018film,
  title={Film: Visual reasoning with a general conditioning layer},
  author={Perez, Ethan and Strub, Florian and De Vries, Harm and Dumoulin, Vincent and Courville, Aaron},
  booktitle={Proceedings of the AAAI conference on artificial intelligence},
  volume={32},
  number={1},
  year={2018}
}

@article{hansen2023idql,
  title={Idql: Implicit q-learning as an actor-critic method with diffusion policies},
  author={Hansen-Estruch, Philippe and Kostrikov, Ilya and Janner, Michael and Kuba, Jakub Grudzien and Levine, Sergey},
  journal={arXiv preprint arXiv:2304.10573},
  year={2023}
}

@article{zhu2024madiff,
  title={Madiff: Offline multi-agent learning with diffusion models},
  author={Zhu, Zhengbang and Liu, Minghuan and Mao, Liyuan and Kang, Bingyi and Xu, Minkai and Yu, Yong and Ermon, Stefano and Zhang, Weinan},
  journal={Advances in Neural Information Processing Systems},
  volume={37},
  pages={4177--4206},
  year={2024}
}

@inproceedings{sohl2015deep,
  title={Deep unsupervised learning using nonequilibrium thermodynamics},
  author={Sohl-Dickstein, Jascha and Weiss, Eric and Maheswaranathan, Niru and Ganguli, Surya},
  booktitle={International conference on machine learning},
  pages={2256--2265},
  year={2015},
  organization={pmlr}
}

@article{song2019generative,
  title={Generative modeling by estimating gradients of the data distribution},
  author={Song, Yang and Ermon, Stefano},
  journal={Advances in neural information processing systems},
  volume={32},
  year={2019}
}

@misc{kingma2013auto,
  title={Auto-encoding variational bayes},
  author={Kingma, Diederik P and Welling, Max and others},
  year={2013},
  publisher={Banff, Canada}
}

@article{gozalo2023chatgpt,
  title={ChatGPT is not all you need. A State of the Art Review of large Generative AI models},
  author={Gozalo-Brizuela, Roberto and Garrido-Merchan, Eduardo C},
  journal={arXiv preprint arXiv:2301.04655},
  year={2023}
}

@article{jo2023promise,
  title={The promise and peril of generative AI},
  author={Jo, A},
  journal={Nature},
  volume={614},
  number={1},
  pages={214--216},
  year={2023}
}

@article{brynjolfsson2025generative,
  title={Generative AI at work},
  author={Brynjolfsson, Erik and Li, Danielle and Raymond, Lindsey},
  journal={The Quarterly Journal of Economics},
  pages={qjae044},
  year={2025},
  publisher={Oxford University Press}
}

@article{goodfellow2020generative,
  title={Generative adversarial networks},
  author={Goodfellow, Ian and Pouget-Abadie, Jean and Mirza, Mehdi and Xu, Bing and Warde-Farley, David and Ozair, Sherjil and Courville, Aaron and Bengio, Yoshua},
  journal={Communications of the ACM},
  volume={63},
  number={11},
  pages={139--144},
  year={2020},
  publisher={ACM New York, NY, USA}
}

@article{ho2016generative,
  title={Generative adversarial imitation learning},
  author={Ho, Jonathan and Ermon, Stefano},
  journal={Advances in neural information processing systems},
  volume={29},
  year={2016}
}

@article{mandlekar2020learning,
  title={Learning to generalize across long-horizon tasks from human demonstrations},
  author={Mandlekar, Ajay and Xu, Danfei and Mart{\'\i}n-Mart{\'\i}n, Roberto and Savarese, Silvio and Fei-Fei, Li},
  journal={arXiv preprint arXiv:2003.06085},
  year={2020}
}

@article{mees2022matters,
  title={What matters in language conditioned robotic imitation learning over unstructured data},
  author={Mees, Oier and Hermann, Lukas and Burgard, Wolfram},
  journal={IEEE Robotics and Automation Letters},
  volume={7},
  number={4},
  pages={11205--11212},
  year={2022},
  publisher={IEEE}
}

@inproceedings{lynch2020learning,
  title={Learning latent plans from play},
  author={Lynch, Corey and Khansari, Mohi and Xiao, Ted and Kumar, Vikash and Tompson, Jonathan and Levine, Sergey and Sermanet, Pierre},
  booktitle={Conference on robot learning},
  pages={1113--1132},
  year={2020},
  organization={Pmlr}
}

@inproceedings{nichol2021improved,
  title={Improved denoising diffusion probabilistic models},
  author={Nichol, Alexander Quinn and Dhariwal, Prafulla},
  booktitle={International conference on machine learning},
  pages={8162--8171},
  year={2021},
  organization={PMLR}
}

@article{vahdat2021score,
  title={Score-based generative modeling in latent space},
  author={Vahdat, Arash and Kreis, Karsten and Kautz, Jan},
  journal={Advances in neural information processing systems},
  volume={34},
  pages={11287--11302},
  year={2021}
}

@article{liang2023adaptdiffuser,
  title={Adaptdiffuser: Diffusion models as adaptive self-evolving planners},
  author={Liang, Zhixuan and Mu, Yao and Ding, Mingyu and Ni, Fei and Tomizuka, Masayoshi and Luo, Ping},
  journal={arXiv preprint arXiv:2302.01877},
  year={2023}
}

@inproceedings{liang2024skilldiffuser,
  title={Skilldiffuser: Interpretable hierarchical planning via skill abstractions in diffusion-based task execution},
  author={Liang, Zhixuan and Mu, Yao and Ma, Hengbo and Tomizuka, Masayoshi and Ding, Mingyu and Luo, Ping},
  booktitle={Proceedings of the IEEE/CVF Conference on Computer Vision and Pattern Recognition},
  pages={16467--16476},
  year={2024}
}

@article{kang2023efficient,
  title={Efficient diffusion policies for offline reinforcement learning},
  author={Kang, Bingyi and Ma, Xiao and Du, Chao and Pang, Tianyu and Yan, Shuicheng},
  journal={Advances in Neural Information Processing Systems},
  volume={36},
  pages={67195--67212},
  year={2023}
}

@article{zhang2024motiondiffuse,
  title={Motiondiffuse: Text-driven human motion generation with diffusion model},
  author={Zhang, Mingyuan and Cai, Zhongang and Pan, Liang and Hong, Fangzhou and Guo, Xinying and Yang, Lei and Liu, Ziwei},
  journal={IEEE transactions on pattern analysis and machine intelligence},
  volume={46},
  number={6},
  pages={4115--4128},
  year={2024},
  publisher={IEEE}
}

@inproceedings{taget,
  title = {Ad Hoc Teamwork via Offline Goal-Based Decision Transformers.},
  author = {Zhang, Xinzhi and Chan, Hohei and Ye, Deheng and Cai, Yi and Zhao{*}, Mengchen},
  year = {2025},
  booktitle = {Proceedings of the 42nd International Conference on Machine Learning},
}

@article{yucesoy2025role,
  title={The role of drones in disaster response: A literature review of operations research applications},
  author={Yucesoy, Ecem and Balcik, Burcu and Coban, Elvin},
  journal={International Transactions in Operational Research},
  volume={32},
  number={2},
  pages={545--589},
  year={2025},
  publisher={Wiley Online Library}
}

@article{hafner2023mastering,
  title={Mastering diverse domains through world models},
  author={Hafner, Danijar and Pasukonis, Jurgis and Ba, Jimmy and Lillicrap, Timothy},
  journal={arXiv preprint arXiv:2301.04104},
  year={2023}
}

@inproceedings{haarnoja2018soft,
  title={Soft actor-critic: Off-policy maximum entropy deep reinforcement learning with a stochastic actor},
  author={Haarnoja, Tuomas and Zhou, Aurick and Abbeel, Pieter and Levine, Sergey},
  booktitle={International conference on machine learning},
  pages={1861--1870},
  year={2018},
  organization={Pmlr}
}

@inproceedings{
zhang2025ad,
title={Ad Hoc Teamwork via Offline Goal-Based Decision Transformers},
author={Xinzhi Zhang and Hohei Chan and Deheng Ye and Yi Cai and Mengchen Zhao},
booktitle={Forty-second International Conference on Machine Learning},
year={2025},
url={https://openreview.net/forum?id=tl3FlgWScA}
}

% \input{./ReproducibilityChecklist.tex}

\appendix

\newpage

\label{sec:app}
\section{Technical Appendices and Supplementary Material}
\subsection{Algorithm pseudocode}
Algorithm \ref{alg:training} demonstrates the training procedure for PADiff. It involves sampling random timesteps, performing forward diffusion, reverse denoising, and computing the loss at each timestep. The total loss is accumulated, and model parameters are updated using gradient descent over multiple epochs.
\vspace{-1em}
\begin{algorithm}[H]
\caption{Training Procedure of PADiff}
\begin{algorithmic}[1]
\label{alg:training}
\STATE Initialize parameters $\theta,\phi,\varphi$
\FOR{Epoch = 1 to N}
    \FOR{each batch}
        \STATE Sample a random timesteps list $l$
        \STATE $L_{\text{total}} \leftarrow 0 $
        \FOR{t in $l$}
        \STATE Sample $k \sim Uniform(\{ 1, \dots, K \})$ for each tuple
        \STATE Perform the forward diffusion for using Eq.5 to get $\bm{a}_t^k$
        \STATE Perform the denoise process using Eq.3 to predict $\bm{a}_t^{k-1}$ and get $h_t^k$
        \STATE Compute Diffusion Loss using Eq.6
        \STATE Predict $\hat{R}_t = R_{\phi}(h^k_t, s_t)$
        \STATE Predict $\hat{G}_t = G_{\varphi}(\hat{R}_t, s_t)$ 
        \STATE Compute $L_{\text{CoReturn}}$ and $L_{\text{CoGoal}}$ using Eq.10 and Eq.11
        \STATE Accumulate total loss $L_{\text{total}} =  L_{\text{Diffusion}} + \alpha L_{\text{CoReturn}} + \beta L_{\text{CoGoal}}$
        \ENDFOR
    \STATE Update parameters $\theta,\phi,\varphi$ using gradient descent
    \ENDFOR
\ENDFOR
\end{algorithmic}
\end{algorithm}

Algorithm \ref{alg:inference} describes the inference process for PADiff. It involves sampling a random action at each timestep and performing denoising iteratively to refine the action. The final action is then executed at each timestep.

\begin{algorithm}[H]
\caption{Inference Process of PADiff}
\begin{algorithmic}[1]
\label{alg:inference}
\FOR{$t = 1, \dots, T$}
    \STATE Sample a random action $\bm{a}_t^K \sim \mathcal{N}(0,\bm{I})$
    \FOR{$k = K, \dots, 1$}
        \STATE Perform the denoise process using Eq.3 to generate $\bm{a}_t^{k-1}$
    \ENDFOR
    \STATE Execute action $\bm{a}_t^{0}$
\ENDFOR
\end{algorithmic}
\end{algorithm}

\begin{table*}[htb]
\centering
\begin{tabular}{lccccccc}
\hline
Methods & PP-4 & LBF-4 & Overcooked-4 & PP-8 & LBF-8 & Overcooked-8 \\
\hline
Diffusion-BC & 53.8 $\pm$ 1.1 & 0.010 $\pm$ 0.005 & 0.24 $\pm$ 0.07 & 57.0 $\pm$ 1.1 & 0.010 $\pm$ 0.006 & 0.24 $\pm$ 0.06\\
Diffusion-QL & 53.7 $\pm$ 1.1 & 0.015 $\pm$ 0.007 & 0.30 $\pm$ 0.08 & 56.7 $\pm$ 1.0 & 0.025 $\pm$ 0.008 & 0.36 $\pm$ 0.08\\
MADiff & 56.3 $\pm$ 1.1 & 0.015 $\pm$ 0.008 & 0.06 $\pm$ 0.03 & 57.8 $\pm$ 1.2 & 0.015 $\pm$ 0.007 & 0.42 $\pm$ 0.08\\
MADT & 57.8 $\pm$ 0.9 & 0.010 $\pm$ 0.006 & 0.06 $\pm$ 0.03 & 58.2 $\pm$ 1.0 & 0.010 $\pm$ 0.006 & 0.12 $\pm$ 0.04\\
CQL & 60.0 $\pm$ 0.9 & 0.020 $\pm$ 0.008 & 0.64 $\pm$ 0.05 & 59.5 $\pm$ 0.9 & 0.000 $\pm$ 0.000 & 0.46 $\pm$ 0.05\\
ODITS & 53.6 $\pm$ 1.0 & 0.080 $\pm$ 0.024 & 0.18 $\pm$ 0.04 & 53.1 $\pm$ 1.2 & 0.000 $\pm$ 0.000  & 0.06 $\pm$ 0.03\\
LIAM & 55.7 $\pm$ 1.1 & 0.000 $\pm$ 0.000 & 0.22 $\pm$ 0.03 & 53.8 $\pm$ 1.1 & 0.065 $\pm$ 0.012 & 0.18 $\pm$ 0.06\\
TAGET & 61.2 $\pm$ 1.0 & 0.080 $\pm$ 0.013 & 0.54 $\pm$ 0.09 & 60.1 $\pm$ 1.1 & 0.065 $\pm$ 0.012 & 0.63 $\pm$ 0.11\\
\textbf{PADiff} & \makecell{\textbf{67.0 $\pm$ 1.1} \\ \textcolor{red}{+9.47\%}} & \makecell{\textbf{0.130 $\pm$ 0.022} \\ \textcolor{red}{+62.50\%}} & \makecell{\textbf{0.88 $\pm$ 0.06} \\ \textcolor{red}{+37.50\%}}& \makecell{\textbf{65.7 $\pm$ 1.0} \\ \textcolor{red}{+9.32\%}}& \makecell{\textbf{0.117 $\pm$ 0.018} \\ \textcolor{red}{+80.00\%} } & \makecell{\textbf{0.71 $\pm$ 0.06} \\ \textcolor{red}{+12.70\%} }\\
\hline
\end{tabular}
\caption{Average Return Comparison with Baselines}
\label{tab:baseline-comparison}
\end{table*}

\subsection{Cross-play Experiments Analysis}
\label{sec:diversity}
In this paper, we use SVD to maximize the differences in value estimates across observation-action pairs between different teams. This design ensures diversity between training and testing policy sets. We conducted the cross-play experiments between different populations, demonstrating significantly lower cooperation efficiency between populations compared to within-population cooperation, empirically validating that the policies represent truly distinct strategies. In cross-population cooperation experiments, each population sequentially designates one agent as the ego agent to collaborate with members from other populations. For example, the matrix notation (Row1, Col4) specifically denotes an experimental configuration where Population 4's designated ego agent interacts with teammates from Population 1, establishing a systematic evaluation framework for inter-population coordination capabilities. As shown in the cross-play matrix in the main body, in each row of the matrix, the diagonal positions behave best, which means that policies between populations don't work well in coordinating, validating the diversity of our test teammate sets.

\begin{figure*}[ht]
\begin{center}
\centerline{\includegraphics[scale=0.7]{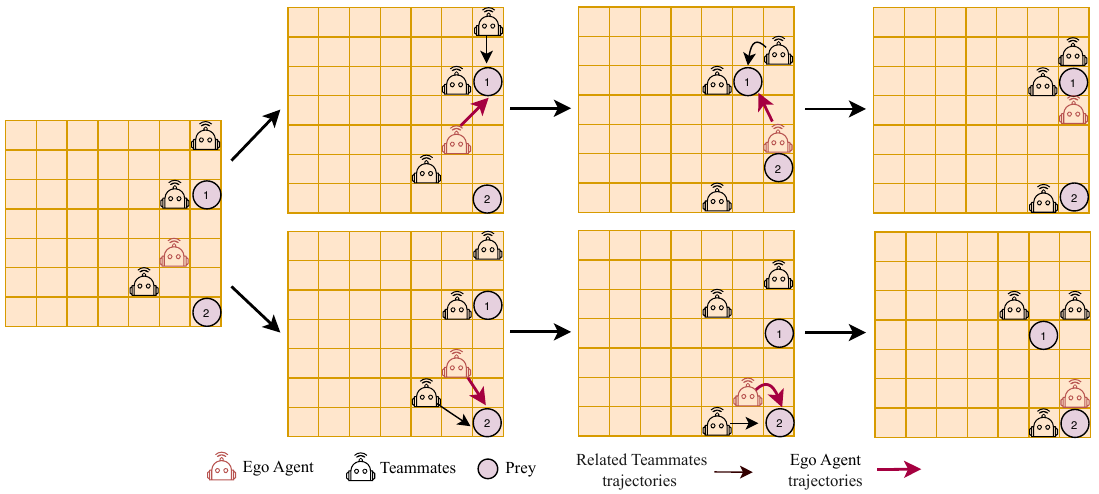}}
    \caption{The visualization of a real scenario in the PP game starting from the same initial state. Our policy has the ability to capture multimodal policy distributions, the ego agent chooses different actions while facing the same state, leading to two different cooperative behaviors.
    } 
    \label{fig:multi-scenario}  
\end{center}

\end{figure*}

\subsection{Multimodal Policy Distributions Analysis}
\label{sec:multi}
\begin{figure}[ht]
\begin{center}
\centerline{\includegraphics[scale=0.032]{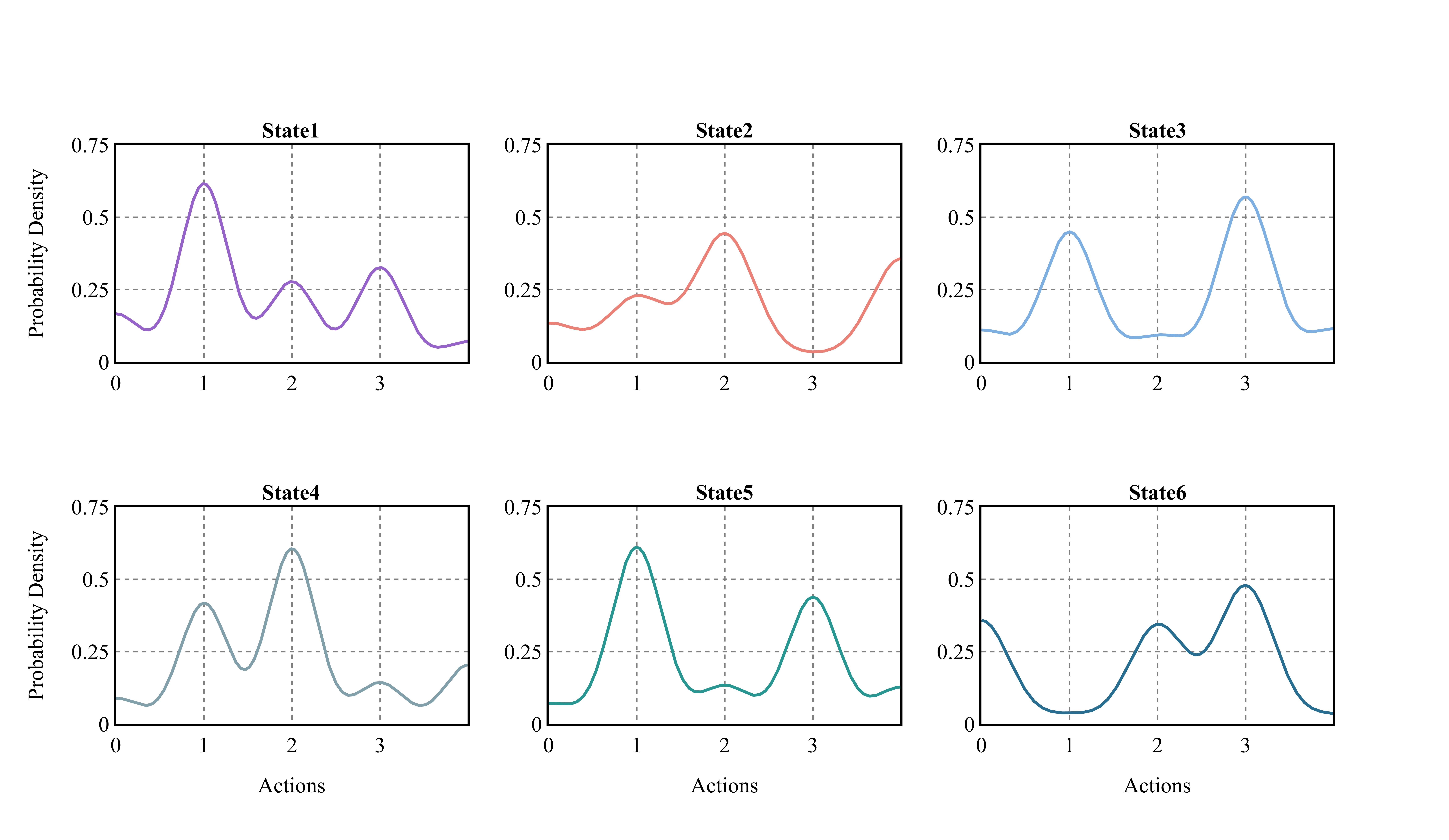}}
    \caption{The distribution visualization of the policy we trained. The probability distribution of the policy was drawn by using the method of kernel probability density estimation, demonstrating our method can fit the policy of a multimodal distribution.
    } 
    \label{fig:multi-plot}  
\end{center}
\vspace{-2em}
\end{figure}

To further demonstrate our model’s ability to capture multimodal policy distributions, we randomly sampled six states from the Predator-Prey (PP) environment. Each state was used as a conditioning input to the model, which was input 1,000 times to generate action samples. As shown in the Figure \ref{fig:multi-plot}, the resulting action distributions exhibit multiple distinct peaks for each state. This indicates that the model successfully learns multimodal action policies.\\

\vspace{-1em}

\subsection{Hyperparameters Analysis}
\label{sec:hyper-analysis}
Our decision‑making model is based on a diffusion process, which generates data by progressively adding noise and learning the denoising steps during training. The number of diffusion steps directly affects the fidelity and quality of the generated outputs. Moreover, in AHT task settings, generalization and robustness are critical. By randomly dropping a proportion of neurons during training, Dropout prevents over‑reliance on specific units and thus enhances both generalization and resilience.\\
To identify optimal parameters for each environment, we conducted a grid of experiments varying both the number of diffusion steps and the dropout rate. Although larger numbers of diffusion steps are often assumed to yield better performance, we observed diminishing returns beyond 30 steps in our domains—while computational cost continued to climb, gains in task performance plateaued. Accordingly, we evaluated diffusion depths of 2, 10, 20, and 30 steps. Our results were shown in Fig\ref{fig:hyper}, demonstrating that a moderate dropout rate combined with an appropriately chosen diffusion depth strikes a favorable balance between training efficiency and resource expenditure, yielding strong performance in AHT scenarios.

\begin{figure*}[t]
\begin{center}
\vspace{-5pt}
\centerline{\includegraphics[scale=0.45]{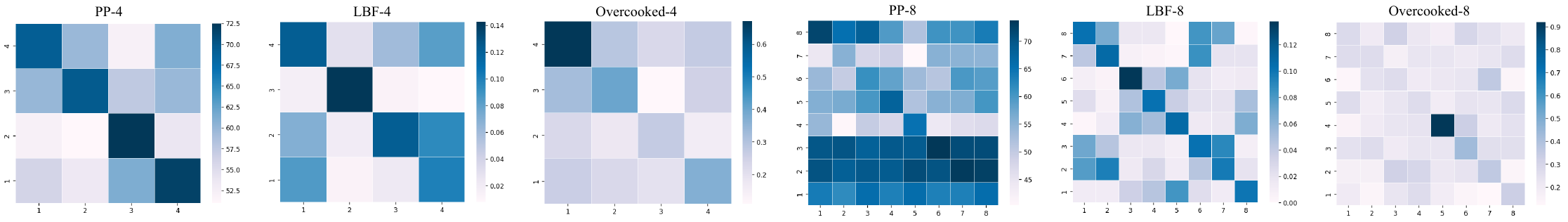}}
    \caption{Cross-play matrices of our testing teammate sets} 
    \label{fig:cross}  
\end{center}
\vspace{-10mm}
\end{figure*}

\begin{figure*}[ht]
\begin{center}
\centerline{\includegraphics[scale=0.045]{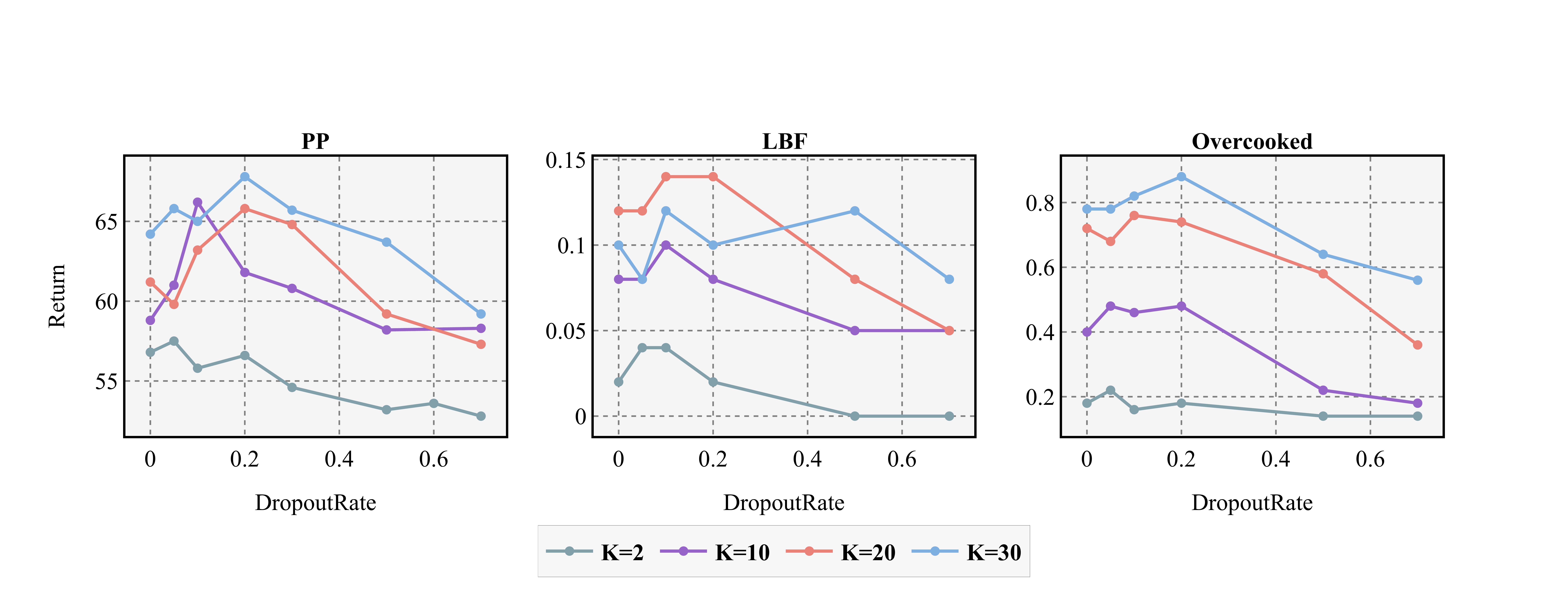}}
    \caption{Comparison among different hyper-parameters
    } 
    \label{fig:hyper}  
\end{center}
\vspace{-3em}
\end{figure*}

\subsection{Computing Resources}
We utilized Quadro RTX 8000 GPUs with 48GB of memory. The batch size for the experiments was set to 128. The execution time for different environments varied: PP environment required 16 hours, LBF took 6 hours, and Overcooked needed 13 hours approximately. This level of detail helps other researchers replicate our experiments under similar conditions, though some variations in resources or setup may still occur.

\subsection{Limitations}
\label{sec:lim}
While our work focuses on improving adaptability and multimodal cooperation in ad hoc teamwork, it still operates under the assumption of full observability, without explicitly modeling uncertainty or partial observability. In the future, we plan to extend our framework to more challenging settings by incorporating mechanisms for handling environmental uncertainty.

\subsection{Networks Details}
\label{sec:net}
\begin{figure*}[ht]
\begin{center}
\centerline{\includegraphics[scale=0.7]{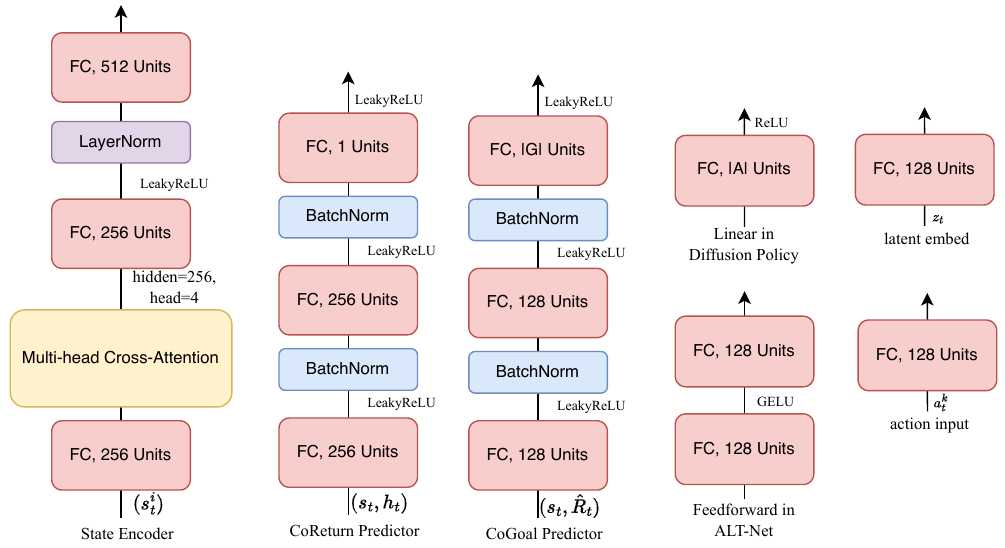}}
    \caption{Architectural details of PADiff
    } 
    \label{fig:net} 
\end{center}
\end{figure*}

As depicted in Figure \ref{fig:net}, we provide full architectural specifications for all modules.

\clearpage

\end{document}